\documentclass[11pt,a4paper]{article}
\usepackage[hyperref]{emnlp2020}
\usepackage{times}
\usepackage{latexsym}

\usepackage{microtype}

\usepackage{times}
\usepackage{latexsym}
\usepackage{url}
\usepackage{times}
\usepackage{latexsym}
\usepackage{amsmath}
\usepackage{amssymb}
\usepackage{amsfonts}
\usepackage{booktabs}
\usepackage{enumitem}
\usepackage{graphicx}
\usepackage{hyperref}
\usepackage{url}
\usepackage{tikz}
\usepackage{xcolor}
\usepackage{pifont}
\usepackage{placeins}
\usepackage[utf8x]{inputenc}
\usepackage{todonotes}
\usepackage{wrapfig}
\usepackage{natbib}
\usepackage{subcaption}
\usepackage{lipsum} 
\usepackage{dblfloatfix}    
\usepackage{float}

\newcommand{\blk}[1]{\textcolor{black}{#1}}
\newcommand{\blu}[1]{\textcolor{blue}{#1}}
\newcommand{\ul}[1]{\underline{#1}}
\newcommand{\gray}[1]{{\color{gray}#1}}
\newcommand{\black}[1]{{\color{black}#1}}
\newcommand{\xv}{\mathbf{x}}
\newcommand{\yv}{\mathbf{y}}

\newcommand{\ourmodel}{Multichannel Generative Language Model} 
\newcommand{\ourtask}{multichannel generative language modeling}
\newcommand{\modelabbv}{MGLM} 
\DeclareMathOperator*{\argmax}{argmax}

\usepackage{amsmath,amsfonts,bm}

\def\eqref#1{equation~\ref{#1}}

\def\1{\bm{1}}

\DeclareMathAlphabet{\mathsfit}{\encodingdefault}{\sfdefault}{m}{sl}
\SetMathAlphabet{\mathsfit}{bold}{\encodingdefault}{\sfdefault}{bx}{n}

\aclfinalcopy 
\def\aclpaperid{3148} 

\title{Multichannel Generative Language Model: \\ Learning All Possible Factorizations Within and Across Channels}

\author{Harris Chan\thanks{\;Work done during an internship at Google Brain.} \\
  Vector Institute \\
  University of Toronto \\
  {\tt hchan@cs.toronto.edu} \\\AND
  Jamie Kiros \\
  Google Research, Brain Team \\
  {\tt kiros@google.com} \\\And
  William Chan \\
  Google Research, Brain Team \\
  {\tt williamchan@google.com}}
\date{}

\begin{document}
\maketitle

\begin{abstract}
A channel corresponds to a viewpoint or transformation of an underlying meaning. A pair of parallel sentences in English and French express the same underlying meaning, but through two separate channels corresponding to their languages.
In this work, we present the Multichannel Generative Language Model (MGLM). MGLM is a generative joint distribution model over channels. MGLM marginalizes over all possible factorizations within and across all channels. 
MGLM endows flexible inference, including unconditional generation, conditional generation (where 1 channel is observed and other channels are generated), and partially observed generation (where incomplete observations are spread across all the channels). 
We experiment with the Multi30K dataset containing English, French, Czech, and German.
We demonstrate experiments with unconditional, conditional, and partially conditional generation. We provide qualitative samples sampled unconditionally from the generative joint distribution.
We also quantitatively analyze the quality-diversity trade-offs and find MGLM outperforms traditional bilingual discriminative models.
\end{abstract}

\section{Introduction}

\begin{figure*}[h!]
    \centering
    \includegraphics[width=\textwidth]{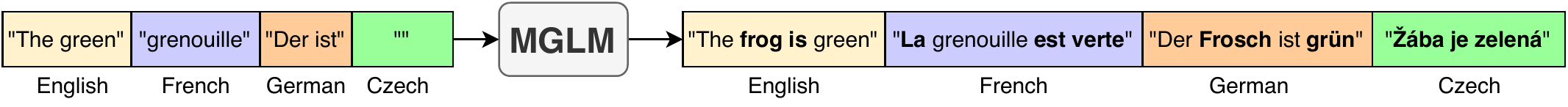}
    \caption{\textbf{Multichannel Generative Language Models} (\modelabbv) marginalize over all possible factorizations of the joint distribution within and across all channels (e.g., languages). 
    \modelabbv{} is trained to predict the tokens to be inserted (in bold), given partially observed inputs.
    At inference, \modelabbv{} can take \textit{full}, \textit{partial}, or \textit{empty} sequence from each channel and generate the full sequence for each channel.}
    \label{fig:mglm_concept}
    \vspace{-12pt}
\end{figure*}

A natural way to consider two parallel sentences in different languages is that each language expresses the same underlying meaning from a different viewpoint. 
Each language can be thought of as a transformation that maps an underlying concept into a view that we collectively agree is determined as `English' or `French'. 
Similarly, an image of a cat and the word `cat' are expressing two views of the same underlying concept. 
In this case, the image corresponds to a high bandwidth channel and the word `cat' to a low bandwidth channel. 
This way of conceptualizing parallel viewpoints naturally leads to the formulation of a fully generative model over each instance, where the transformation corresponds to a particular generation of the underlying view. 
We define each of these views as a channel. As a concrete example, given a parallel corpus of English and French sentences, English and French become two channels, and the corresponding generative model becomes $p(\mathrm{English}, \mathrm{French})$. 
One key advantage of this formulation is that a single model can be trained to capture the full expressivity of the underlying concept, allowing us to compute conditionals and marginals along with the joint. 
In parallel sentences, the conditionals correspond to translations from one channel to another while the marginals correspond to standard monolingual language models.

In this work, we present a general framework for modeling the joint distribution $p(\xv_1,...,\xv_k)$ over $k$ channels by marginalizing over all possible factorizations across the channels and within each channel. 
This formulation allows our framework to perform: 1) unconditional generation, 2) fully conditional generation (source channels are fully observed and fixed), and 3) \textit{partial} conditional generation (source channels contain incomplete sequences).

The key contributions in this work are:
\begin{enumerate}
    \item We present \modelabbv{}, a multichannel generative modeling framework. \modelabbv{} models the joint distribution $p(\xv_1, \dots, \xv_k)$ over $k$ channels by marginalizing over all possible factorization across and within sequences.
    \item Since \modelabbv{} is trained over all possible factorizations, \modelabbv{} can perform both conditional generation (e.g., machine translation with fully observed source channel), and \textit{partially observed} conditional generation across different channels (e.g., seeding each channel with different words, and sample sentences consistent with each other).
    \item In the case of conditional generation over multiple target languages, we show that we are competitive in BLEU and have significant advantages in inference time and model memory savings.
    \item We analyze the Quality-Diversity tradeoff from sampling \modelabbv{} and prior work.
\end{enumerate}

We highlight that while we focus on languages as a specific instantiation of a channel, our framework can generalize to any arbitrary specification, such as other types of tasks (e.g., question-answering) or other modalities (e.g., image captioning). 

\section{\ourmodel{}}
\begin{figure*}[t]
    \centering
    \begin{subfigure}{0.48\textwidth}
        \centering
        \includegraphics[width=\textwidth]{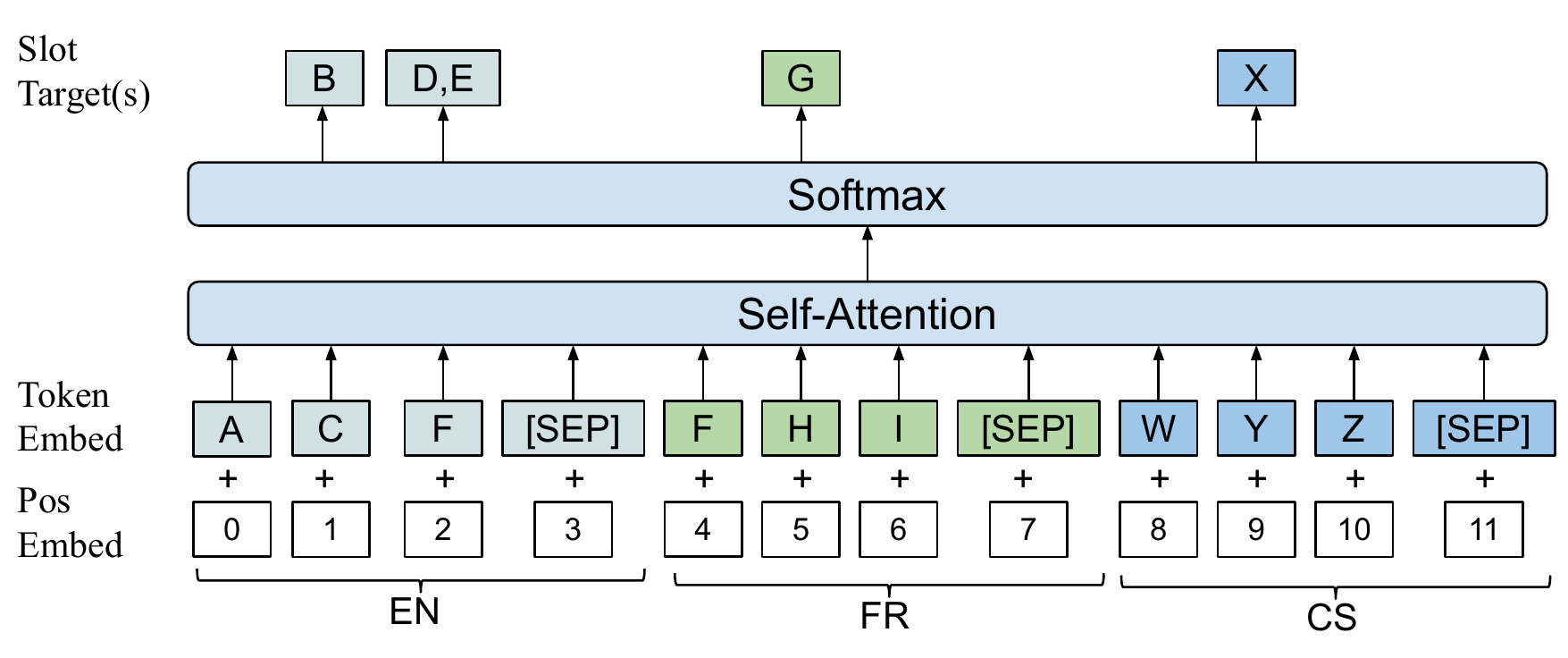}
        \caption{MGLMs Training}
        \label{fig:model_train}
    \end{subfigure}
    \begin{subfigure}{0.48\textwidth}
        \centering
        \includegraphics[width=\textwidth]{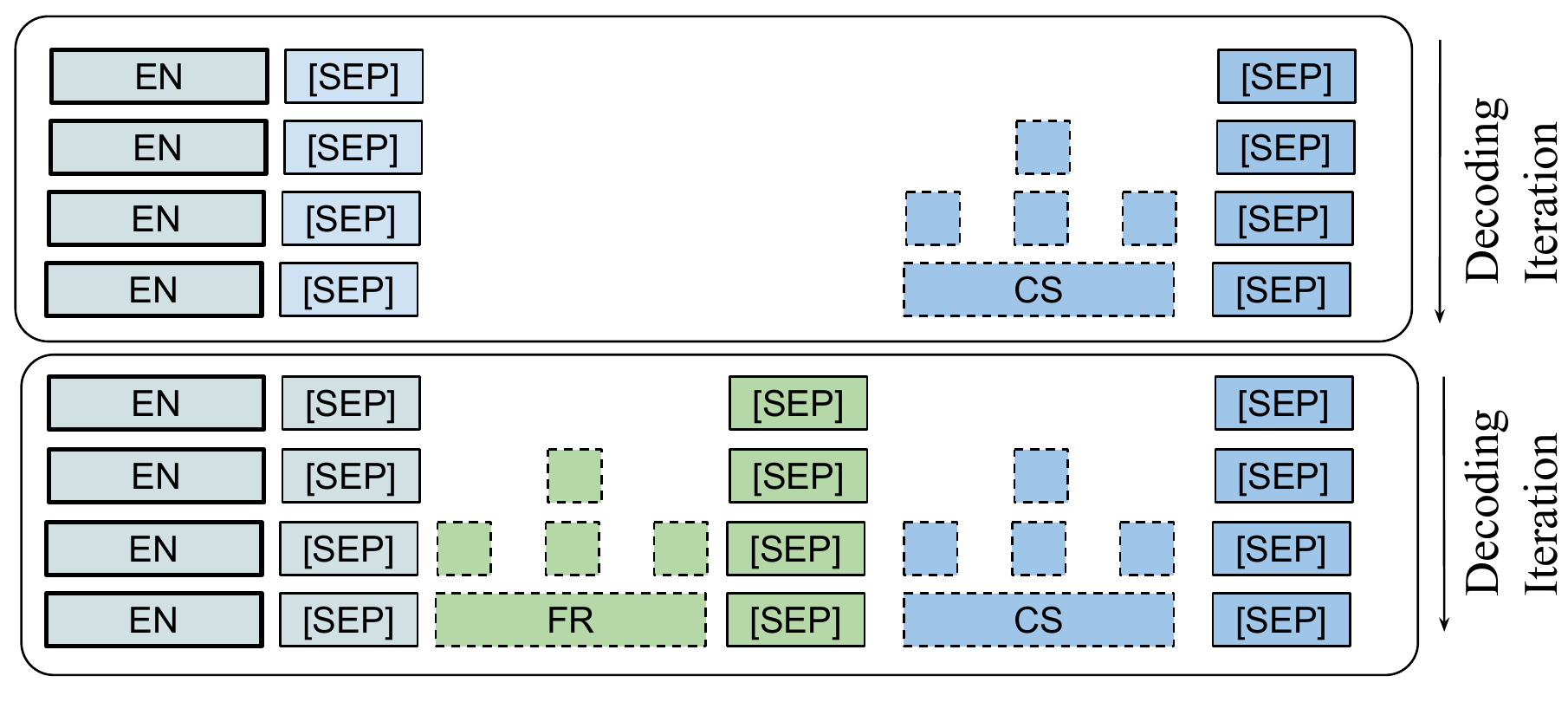}
        \caption{MGLMs Inference}
        \label{fig:model_inference}
    \end{subfigure}
    \caption{(\subref{fig:model_train}) An example multichannel modeling over 3 languages (English, French, Czech), where the model predicts the missing tokens at each location across multiple channels. (\subref{fig:model_inference}) During inference, \modelabbv{} \ can generate output sequence for a single target language channel (top) or multiple language channels in parallel (bottom), conditioning on source channel sentence, and partial translations of multiple language channels.}
    \label{fig:data_subset}
\end{figure*}

In \ourtask{}, our goal is to learn a generative model given a dataset consisting of a set of sequences $\{\xv_1^{(i)}, \dots, \xv_k^{(i)}\}_{i=1}^M$ from up to $k$ channels, where $\xv^{(i)}_j = [x^{(i)}_{j,1}, \dots, x^{(i)}_{j,n}]$ represents a sequence of tokens from the $j$-th channel for the $i$-th example. The \modelabbv{} \ models a joint generative distribution over multiple channels: $p(\xv_1, \dots, \xv_k)$ using all possible factorizations.

\subsection{Joint Probability Factorization}
Given multiple sequences, each from different channels, there are many possible ways to factorize the joint probability over the channels. One approach to treat the channels as a \textit{sequence} of channels, and use an autoregressive left-to-right model over the sequence of channels:
\begin{align*}
    p(\xv_1, \dots, \xv_k) = p(\xv_1)\prod_i p(\xv_i | \xv_1,...,\xv_{i-1})
\end{align*}
Within each channel, the token sequence probability can also be modeled autoregressively:
\begin{align*}
    p(\xv_i | \xv_1,...\xv_{i-1}) = \prod_t p(x_{i,t} | \xv_1,...,\xv_{i-1},\xv_{i,<t})
\end{align*}
This approach assumes: (1) a particular ordering over the channels; (2) the completion sequences from previous channels before generating the next channel's sequence. These assumptions are valid in some applications. For example, bilingual machine translation is a special case where $k=2$, the channels are languages, and the source and target languages dictate the ordering over the channels and its token sequences.

In \modelabbv, we instead consider a more general approach, wherein we marginalize over all possible factorization order. Let $z$ represent the permutation of indices $\{1,\dots,N\}$ where $N$ is the total number of tokens summed across all the channels. The joint probability is marginalized over $z$:
\begin{align} \label{eqn:joint}
    p(\xv_1, \dots, \xv_k) = \sum_{z \in S_N} p(z)p(\xv_1, \dots, \xv_k | z),
\end{align}
Where $p(z)$ denotes a prior over the different possible permutation, which can be uniform or a balanced binary tree prior \citep{stern-icml-2019}. 
Unfortunately, computing the exact log-likelihood in Eqn. \ref{eqn:joint} is intractable due to marginalization over all permutation order $z$. In practice, we optimize its lower bound via Jensen's inequality:
\begin{align}
&\log p(\xv_1, \dots, \xv_k) \nonumber \\
& = \log \sum_{z \in S_N} p(z)p(\xv_1, \dots, \xv_k \mid z) \label{log_likelihood} \\
& \ge \sum_{z \in S_N} p(z) \log p(\xv_1, \dots, \xv_k \mid z) =: \mathcal{L}(\{\xv_i\}^k) \label{eqn:lowerbound_loss}
\end{align}

\subsection{Model Architecture}
One natural class of models for \modelabbv{} is the insertion-based Transformer \citep{stern-icml-2019,welleck-icml-2019,gu-arxiv-2019}, which considers arbitrary factorization of the output sequence by using insertion operation, predicting both (1) content token $c \in \mathcal{C}$ from the vocabulary, and (2) location $l \leq t$ to insert, relative to (e.g. to the left of) the current partial output $\hat{\yv}_t$:
\begin{align}
    p(c,l|\xv,\hat{\yv}_t) = \text{InsertionTransformer}(\xv,\hat{\yv}_t)
\end{align}
The (content, location) distribution is factorized as $p(c,l) = p(c|l)p(l)$, where $p(c|l)$ is the standard Transformer softmax over vocabulary, and $p(l)$ is the softmax over the locations.
KERMIT \citep{chan2019kermit} further simplified the Insertion Transformer model by removing the encoder and only having a decoder stack \citep{vaswani-nips-2017}, by concatenating the original input and output sequence as one single sequence and optimizing over all possible factorizations. Consequently, KERMIT is able to model the joint $p(\xv, \yv)$, conditionals $p(\xv \mid \yv)$, $p(\yv \mid \xv)$, as well as the marginals $p(\xv), p(\yv)$. We extend KERMIT to consider using a Transformer decoder for modeling the joint probability over $k>2$ channels.  

\subsection{Training}
Without the loss of generality, we denote $\xv = [\xv_1,\dots,\xv_k]$ as the concatenation of the $k$ sequences\footnote{The set of permutation $z \in S_N$ includes different order of channels as well}. With the insertion framework, the loss function Eqn. (\ref{eqn:lowerbound_loss}) can be simplified by changing the summation and careful decomposition of the permutation, leading to:
\begin{align*}
\mathcal{L}(\xv) &= \sum_{z \in S_N} p(z) \log \prod_{i=1}^N p((c^z_i, l^z_i) \mid \xv^{z,i-1}_{1:i-1}) \\
& = \sum_{i=1}^N \sum_{z_{1:i-1}} p(z_{1:i-1})\sum_{z_i} p(z_i \mid z_{1:i-1}) \\ 
& \quad \quad \log p((c^z_i, l^z_i) \mid \xv^{z,i-1}_{1:i-1})
\end{align*}
We illustrate an example data input consisting of 3 channels in Figure \ref {fig:model_train}. We concatenate the sequences together from all channels for each example, separated by a \texttt{SEP} token. Even with a shared vocabulary, each channel results in a different token embedding, via addition of a channel-specific (learnable) embedding, or simply having a separately learned token embedding per channel. After passing through the dense self-attention layers as per Transformer architecture, the contextualized representation at each output time step predicts the possible tokens to be inserted to the left of the current input token. For a uniform prior $p(z)$, the target tokens at each slot are weighted equally.  

\subsection{Inference}
At inference (generation) time, we can generate unconditionally by seeding the canvas with the \texttt{[SEP]} token and predicting the first actual token or provide as much, or as little, partial/complete sequence in each channel. Each output token is chosen via sampling or greedily choosing a single (content, location) with maximum probability in the partial canvas $\hat{\xv}_t$:
\begin{align}
    (\hat c, \hat l) = \argmax_{c, l} p(c, l | \hat{\xv}_t),
\end{align}
or inserted in all available insertion slots at once, in parallel:
\begin{align}
    \hat{c}_{l} = \argmax_{c} p(c \mid l, \hat{\xv}_t),
\end{align}
Figure \ref{fig:model_inference} shows two example decoding inference: a single target language channel (top), or multiple target language channels in parallel (bottom). Note that for both cases, each channel inserts in all available slots. 

\section{Related Work}

MGLM was inspired and influenced by prior work on conditional and unconditional language modeling. Insertion Transformer \citep{stern-icml-2019} and XLNet \citep{yang2019xlnet} also marginalize over all possible factorizations. However, their work is focused on the conditional distribution $p(\yv|\xv)$, and they do not marginalize over all possible factorizations of the joint distribution. MGLM can be viewed as an extension and generalization of KERMIT \citep{chan2019kermit}. KERMIT is a generative joint distribution model that also learns all possible factorizations. However, KERMIT is constrained to two languages, while MGLM is a generative joint distribution model across any/all languages/text while learning all possible factorizations of the joint distribution. 

MGLM follows from prior works on cross-lingual language models, which aim to learn shared representation across languages. XLM \citep{conneau2019unsupervised} is closely related to our work and also concatenate source and target sequences from different languages; however, their work is limited to bilingual concatenation, is not fully generative, and requires length conditioning. MGLM is not limited to two languages and generalizes to multiple channels/languages, is fully generative, and our insertion-based approach (as opposed to masking-based approach) does not require length conditioning. Multilingual Neural Language Model \citep{wada2018unsupervised} uses a shared encoder and language-dependent decoders to generate word embeddings and evaluate word alignment tasks. In contrast, our work unifies the neural architecture with a straightforward stack of self-attention layers. Finally, \citet{dong2015multi} explored multi-task learning for machine translation with an autoregressive network. The key difference between our work and other prior work on multi-target or multi-task learning is that MGLM models all possible factorizations of the joint distribution across all channels, instead of just the left-to-right factorization. This difference licenses MGLM to perform any form of sampling (conditional, unconditional, partially-conditional) without any rigid left-to-right restrictions.

Evaluation of text generative models remain a challenge \citep{liu2016not,novikova2017we}. Quality versus diversity plots have been used to compare the trade-off at different output softmax temperatures, as such in Stochastic Beam Search \citep{kool2019stochastic}, which used a simpler $n$-gram diversity instead of Self-BLEU \citep{zhu2018texygen}. However, we are the first to characterize the Q-D behaviour of insertion based models versus existing left-to-right language models. Other metrics summarize the quality and diversity trade-off as a single number, such as Fr\'{e}chet BERT Distance \citep{montahaei2019jointly} inspired by the FID score \citep{heusel2017gans} used in computer vision, or take into account human evaluation \citep{hashimoto2019unifying}.

\section{Experiments}

\begin{table*}[t]
\centering
\small
\begin{tabular}{lcccc}
\toprule
\bfseries Model & Inference & Test2016  & Test2017 & MSCOCO  \\
\midrule
Bilingual (EN $\rightarrow$ DE) & EN $\rightarrow$ DE & 36.14 & 28.32 & 24.15 \\
Bilingual (EN $\leftrightarrow$ DE) & EN $\rightarrow$ DE & \textbf{37.08} & 28.69 & 26.11  \\
\midrule
Multi-target (EN $\rightarrow$ Rest) & EN $\rightarrow$ DE & 36.83 & 28.35 & 25.14 \\
                                     & EN $\rightarrow$ FR,CS,\textbf{DE} & 35.41 & \textbf{29.69} & 25.64 \\
\midrule
Multi-target (Any $\rightarrow$ Rest) & EN $\rightarrow$ DE & 36.63 & 28.37  & \textbf{26.98}  \\
                                     & EN $\rightarrow$ FR,CS,\textbf{DE} & 36.51  & 28.53 & 25.84  \\
\midrule
Joint ($p(EN,FR,CS,DE)$) & EN $\rightarrow$ DE & 33.06 & 23.42 & 21.39 \\
                        & EN $\rightarrow$ FR,CS,\textbf{DE} & 32.53 & 23.78 & 20.97  \\
\bottomrule
\end{tabular}%
\caption{Multi30k English $\rightarrow$ German test BLEU. Higher is better.}
\label{tab:multi30k_en2de}
\end{table*}

We experiment on a multilingual dataset to demonstrate that we can learn  \modelabbv{}. We perform both qualitative and quantitative experiments. We highlight the model's capabilities ranging from conditional generation (i.e., machine translation) to unconditional sampling of the joint distribution over multiple languages.

\begin{figure*}[t]
    \small
    \centering
    \color{gray}\begin{tabular}{p{14cm}}
        \blk{\textbf{EN Input:} A man sits on a bench holding his dog and looking at the water. } \\
        \blk{\textbf{Parallel Decode:}} \\
        \hline
        \blk{\textbf{FR:}} {\_}Un {\_}homme {\_}est {\_}assis {\_}sur {\_}un {\_}banc , {\_}ten \ul{\blu{ant}} {\_}son {\_}chien {\_}et {\_}regardant {\_}l ' eau . \blk{[SEP]} \\
        \blk{\textbf{CS:}}  {\_}Muž {\_}sedí {\_}na {\_}lavičce {\_}a {\_}drží \ul{\blu{{\_}své}} ho {\_}psa {\_}a {\_}dívá {\_}se {\_}na {\_}vodu . \blk{[SEP]} \\
        \blk{\textbf{DE:}} {\_}Ein {\_}Mann {\_}sitzt {\_}auf {\_}einer {\_}Bank {\_}und \ul{\blu{{\_}hält}} {\_}seine n {\_}Hund {\_}und {\_}schaut {\_}auf {\_}das {\_}Wasser . \blk{[SEP]} \\
        \hline
         \blk{\textbf{FR:}}{\_}Un {\_}homme {\_}est {\_}assis \ul{\blu{{\_}sur}} {\_}un {\_}banc , {\_}ten {\blk{ant}} {\_}son {\_}chien {\_}et {\_}regardant \ul{\blu{{\_}l}} ' eau . \blk{[SEP]} \\
         \blk{\textbf{CS:}}{\_}Muž {\_}sedí {\_}na \ul{\blu{{\_}lavičce}} {\_}a {\_}drží {\blk{{\_}své}} ho {\_}psa \ul{\blu{{\_}a}} {\_}dívá {\_}se {\_}na {\_}vodu . \blk{[SEP]} \\
         \blk{\textbf{DE:}} {\_}Ein {\_}Mann {\_}sitzt {\_}auf \ul{\blu{{\_}einer}} {\_}Bank {\_}und {\blk{{\_}hält}} {\_}seine n {\_}Hund \ul{\blu{{\_}und}} {\_}schaut {\_}auf {\_}das {\_}Wasser . \blk{[SEP]} \\
         \hline
         \blk{\textbf{FR:}} {\_}Un {\_}homme \ul{\blu{{\_}est}} {\_}assis {\blk{{\_}sur}} {\_}un \ul{\blu{{\_}banc}} , {\_}ten {\blk{ant}} {\_}son \ul{\blu{{\_}chien}} {\_}et {\_}regardant {\blk{{\_}l}} ' \ul{\blu{eau}} . \blk{[SEP]} \\
         \blk{\textbf{CS:}} {\_}Muž \ul{\blu{{\_}sedí}} {\_}na {\blk{{\_}lavičce}} \ul{\blu{{\_}a}} {\_}drží {\blk{{\_}své}} \ul{\blu{ho}} {\_}psa {\blk{{\_}a}} {\_}dívá {\_}se {\_}na \ul{\blu{{\_}vodu}} . \blk{[SEP]} \\
         \blk{\textbf{DE:}} {\_}Ein {\_}Mann \ul{\blu{{\_}sitzt}} {\_}auf {\blk{{\_}einer}} \ul{\blu{{\_}Bank}} {\_}und {\blk{{\_}hält}} {\_}seine n \ul{\blu{{\_}Hund}} {\blk{{\_}und}} {\_}schaut {\_}auf \ul{\blu{{\_}das}} {\_}Wasser . \blk{[SEP]} \\ 
        \hline
        \blk{\textbf{FR:}} \ul{\blu{{\_}Un}} {\_}homme {\blk{{\_}est}} \ul{\blu{{\_}assis}} {\blk{{\_}sur}} \ul{\blu{{\_}un}} {\blk{{\_}banc}} , \ul{\blu{{\_}ten}} {\blk{ant}} \ul{\blu{{\_}son}} {\blk{{\_}chien}} {\_}et \ul{\blu{{\_}regardant}} {\blk{{\_}l}} \ul{\blu{'}} {\blk{eau}} \ul{\blu{.}} \blk{[SEP]} \\
         \blk{\textbf{CS:}} \ul{\blu{{\_}Muž}} {\blk{{\_}sedí}} \ul{\blu{{\_}na}} {\blk{{\_}lavičce}} {\blk{{\_}a}} \ul{\blu{{\_}drží}} {\blk{{\_}své}} {\blk{ho}} \ul{\blu{{\_}psa}} {\blk{{\_}a}} {\_}dívá \ul{\blu{{\_}se}} {\_}na {\blk{{\_}vodu}} \ul{\blu{.}} \blk{[SEP]} \\
         \blk{\textbf{DE:}} \ul{\blu{{\_}Ein}} {\_}Mann {\blk{{\_}sitzt}} \ul{\blu{{\_}auf}} {\blk{{\_}einer}} {\blk{{\_}Bank}} \ul{\blu{{\_}und}} {\blk{{\_}hält}} {\_}seine \ul{\blu{n}} {\blk{{\_}Hund}} {\blk{{\_}und}} {\_}schaut \ul{\blu{{\_}auf}} {\blk{{\_}das}} {\_}Wasser \ul{\blu{.}} \blk{[SEP]} \\
        \hline
        \blk{\textbf{FR:}} {\blk{{\_}Un}} \ul{\blu{{\_}homme}} {\blk{{\_}est}} {\blk{{\_}assis}} {\blk{{\_}sur}} {\blk{{\_}un}} {\blk{{\_}banc}} \ul{\blu{,}} {\blk{{\_}ten}} {\blk{ant}} {\blk{{\_}son}} {\blk{{\_}chien}} \ul{\blu{{\_}et}} {\blk{{\_}regardant}} {\blk{{\_}l}} {\blk{'}} {\blk{eau}} {\blk{.}} \blk{[SEP]} \\
         \blk{\textbf{CS:}} {\blk{{\_}Muž}} {\blk{{\_}sedí}} {\blk{{\_}na}} {\blk{{\_}lavičce}} {\blk{{\_}a}} {\blk{{\_}drží}} {\blk{{\_}své}} {\blk{ho}} {\blk{{\_}psa}} {\blk{{\_}a}} \ul{\blu{{\_}dívá}} {\blk{{\_}se}} \ul{\blu{{\_}na}} {\blk{{\_}vodu}} {\blk{.}} \blk{[SEP]} \\
         \blk{\textbf{DE:}} {\blk{{\_}Ein}} \ul{\blu{{\_}Mann}} {\blk{{\_}sitzt}} {\blk{{\_}auf}} {\blk{{\_}einer}} {\blk{{\_}Bank}} {\blk{{\_}und}} {\blk{{\_}hält}} \ul{\blu{{\_}seine}} {\blk{n}} {\blk{{\_}Hund}} {\blk{{\_}und}} \ul{\blu{{\_}schaut}} {\blk{{\_}auf}} {\blk{{\_}das}} \ul{\blu{{\_}Wasser}} {\blk{.}} \blk{[SEP]} 
    \end{tabular}
    \color{black}
    \caption{Example parallel greedy decode using the Multi-target (Any $\rightarrow$ Rest) KERMIT model, starting with an English sentence. Blue underlined tokens are the inserted tokens at each iteration, and the gray tokens are the final output tokens that have not been generated yet. The three target languages are generated together in parallel.}
    \label{fig:parallel_decode}
\end{figure*}

We experiment on the Multi30k\footnote{\url{https://github.com/multi30k/dataset}} \citep{elliott2016multi30k,elliott2017findings,barrault2018findings}, a multilingual dataset which consists of 29,000 parallel training sentences in English (EN), French (FR), Czech (CS), and German (DE) sentences. We use Multi30k because multiple high-quality channels (multilingual translations) are readily available to highlight our framework.
We implement \modelabbv{} \ as a base Transformer decoder, without any causal masking, with 6 hidden layers and 1024 dimensional hidden representation. We concatenate all 4 language raw text training examples and use SentencePiece \citep{kudo2018sentencepiece} to learn a universal subword unigram \citep{kudo2018subword} tokenizer with a shared 32K vocabulary size. 
We follow a similar training set up to BERT \citep{devlin-naacl-2019}, using Adam \citep{kingma-iclr-2015} optimizer with a learning rate of 1e-4, warm up over the first 10\% of the total training iterations varying between 10k to 50k iterations. 
We can train 3 different variants of \modelabbv{} by altering the sampling ratio of training data seen by the model:
\begin{enumerate}
    \item \textbf{Bilingual} (e.g., EN $\rightarrow$ FR). We give the model a fully observed source (e.g., $EN$), and ask the model to infill the target (e.g., $FR$).
    \item \textbf{Multi-target} (e.g., any 1 $\rightarrow$ Rest). We give the model a fully observed source (e.g., $EN$), and ask the model to infill the rest of the targets (e.g., $DE$, $FR$, $CS$).
    \item \textbf{Joint}. We ask the model to infill all the targets, consequently we learn a joint distribution over all the languages $p(\mathrm{en}, \mathrm{fr}, \mathrm{de}, \mathrm{cs})$.
\end{enumerate}

\subsection{Translation Performance}
\begin{figure*} [t]
    \centering
    \begin{subfigure}{0.32\textwidth}
        \centering
        \includegraphics[width=0.95\linewidth]{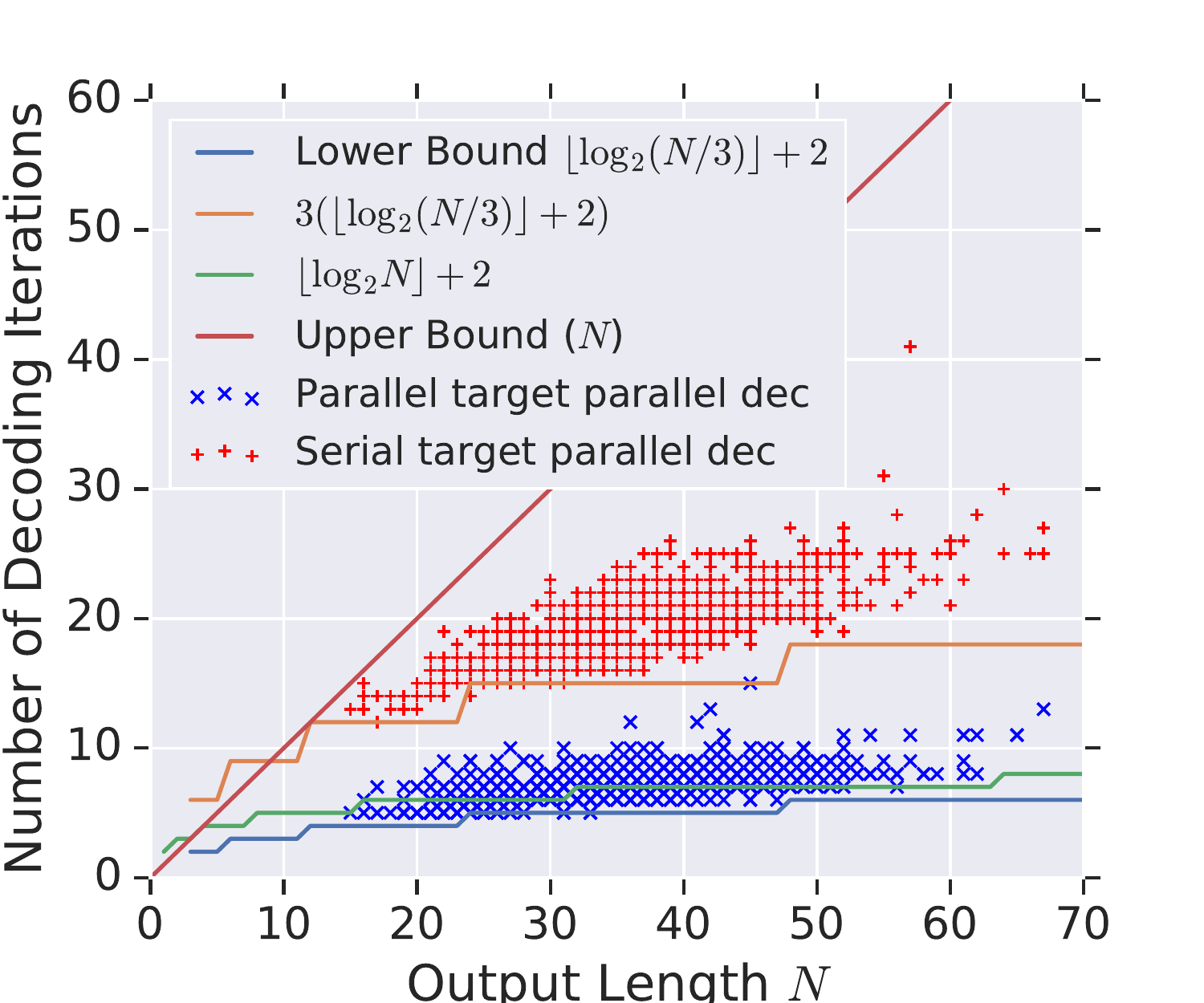}
        \caption{Number of Decoding Iteration}
        \label{fig:par_decit_outlen}
    \end{subfigure}
    \begin{subfigure}{0.32\textwidth}
        \centering
        \includegraphics[width=0.95\linewidth]{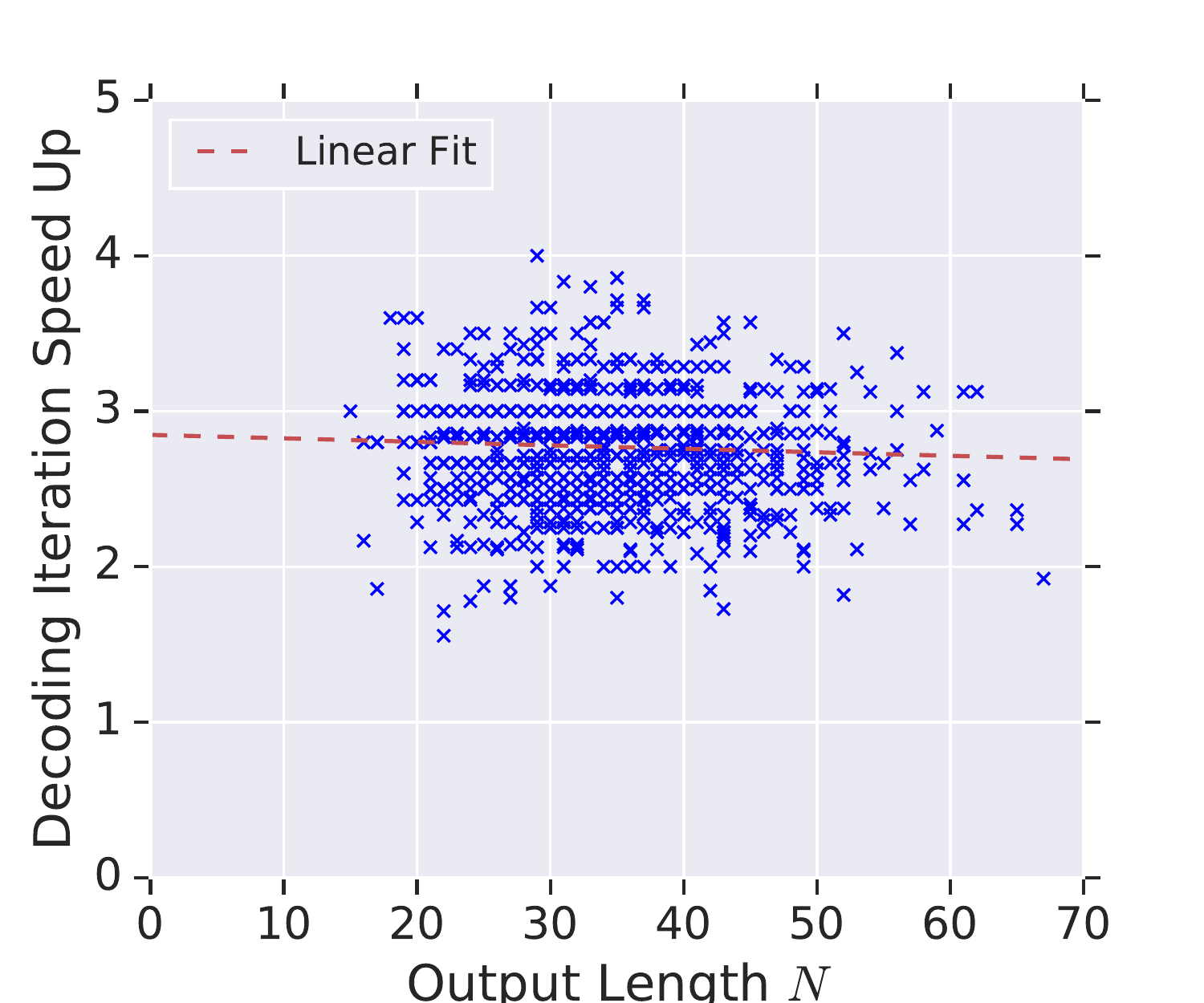}
        \caption{Decoding Iteration Speedup}
        \label{fig:par_decsp_outlen}
    \end{subfigure}
    \begin{subfigure}{0.32\textwidth}
        \centering
        \includegraphics[width=0.95\linewidth]{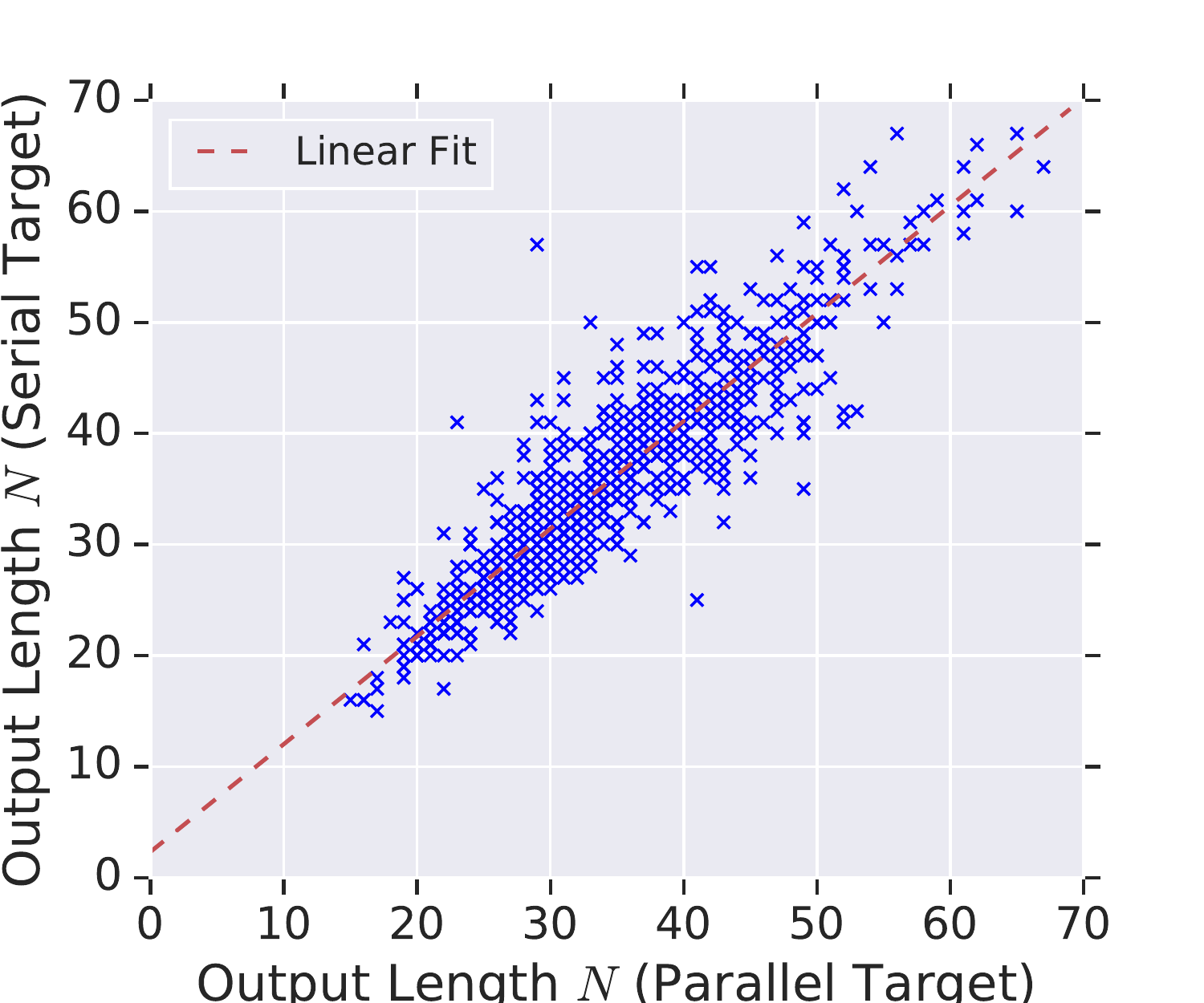}
        \caption{Serial vs. parallel target output length}
        \label{fig:par_outser_outpar}
    \end{subfigure}
    \caption{(\subref{fig:par_decit_outlen}) The number of decoding iterations vs. the output length when decoding each target language serially vs. in parallel, compared to various logarithmic bounds. We have shown that the model can achieve close to the theoretical lower bound $\lfloor \log_2(N/k) \rfloor + 2$ where the number of target languages $k=3$. (\subref{fig:par_decsp_outlen}) Relative wall-clock speed up when using the parallel target languages decoding vs. serial, achieving slightly under 3 times speed up. (\subref{fig:par_outser_outpar}) Total output length for the 3 target languages when using serial vs. parallel target language generation. While not identical, we observe a linear relationship between the output length using the two different modes}
    \label{fig:par_speed_comp}
\end{figure*}

The goal of \modelabbv{} is not conditional generation (i.e., machine translation), but nevertheless, we demonstrate its ability to do conditional generation in this section.
We report the BLEU scores \citep{papineni2002bleu} on the three test sets: \texttt{test 2016 Flickr}, \texttt{test 2017 Flickr}, \texttt{test 2017 MSCOCO}, for different English $\rightarrow$ \{German, French, Czech\} translations. We use parallel greedy decoding \citep{stern-icml-2019,chan2019kermit}, i.e. inserting to all incomplete slots. Table \ref{tab:multi30k_en2de} summarizes the results for English to German. Additional results for English to French, English to Czech, and German to English are shown in Appendix \ref{app:more_multi30k_bleu}. 
We observe that the Multi-target models performed similar to or slightly better than the bilingual models trained only on a single language pair.
This is particularly useful when multiple machine translation targets are desired. We now only need one \modelabbv{}, which is competitive to the bidirectional expert models. This implies we only need 1 model for inference over multiple languages, instead of $N$ models (i.e., saving substantial memory).

We also observe the full generative joint model has a BLEU gap compared to the bilingual baseline, consistent with the findings in \citet{chan2019kermit}. We hypothesize this is due to the joint distribution being a more challenging task. We further hypothesize that the joint model needs to fantasize additional details when conditioning on the partial sequence in each channel during training. This results in fantasizing additional details not present in the source sentence during translation tasks.

\subsection{Parallel Greedy Decoding: Parallel in Target Languages}
As alluded conceptually in Figure \ref{fig:data_subset} and in the previous section, our KERMIT-based \modelabbv{} is also able to perform parallel greedy decoding that is also \textit{parallel in the number of target languages}. 
We illustrate this process in Figure \ref{fig:parallel_decode}. 
By starting with $K$ initial \texttt{[SEP]} tokens for $K$ target output languages, \modelabbv{} can decode $K$ target languages that have at most $n$ output tokens per language in $\mathcal{O}(\log{}n)$, i.e. constant in the number of target languages.

We investigate the relative speed up in generating multiple target language outputs in parallel versus generating the targets in series, in terms of wall-clock time and the number of decoding iterations. In Figure \ref{fig:par_decit_outlen}, we plot the number of decoding iterations taken versus the total output length $N$ for each sentence in the \texttt{test 2016 Flickr} test set, using the Joint \modelabbv{} model when decoding from a single source language to 3 target languages: English $\rightarrow$  \{French, German, Czech\}. When performing serial target decoding, we only output the target conditioned on English, i.e., English $\rightarrow$ French, English $\rightarrow$ German, English $\rightarrow$ Czech. We also plot several theoretical bounds: (1) upper bound ($N$) when decoding entirely serially, (2) lower bound $3(\lfloor \log_2(N/3) \rfloor + 2)$ when decoding 3 languages serially but parallel within each language, (3) lower bound $\lfloor \log_2(N/3) \rfloor + 2$, when decoding the 3 target languages in parallel and parallel within each language, and (4) $\lfloor \log_2(N) \rfloor + 2$, if we decode the entire output in parallel as a single sequence. We observe that our model can meet the lower bound several times and often decode below the fourth $\lfloor \log_2(N) \rfloor + 2$ bound. Figure \ref{fig:par_decsp_outlen} compares the wall-clock speed up when decoding targets in parallel vs. in series, with a linear regression line plotted. Our model achieving almost 3 times speedup in wall-clock speed. The parallel targets decoding is bottlenecked by the target language with the longest output sequence. Figure \ref{fig:par_outser_outpar} compares the total output length when decoding the targets in series versus in parallel. We observe that there is a linear relationship between the output lengths using the two modes.

\begin{figure*}[t]
    \centering
    \includegraphics[width=\textwidth]{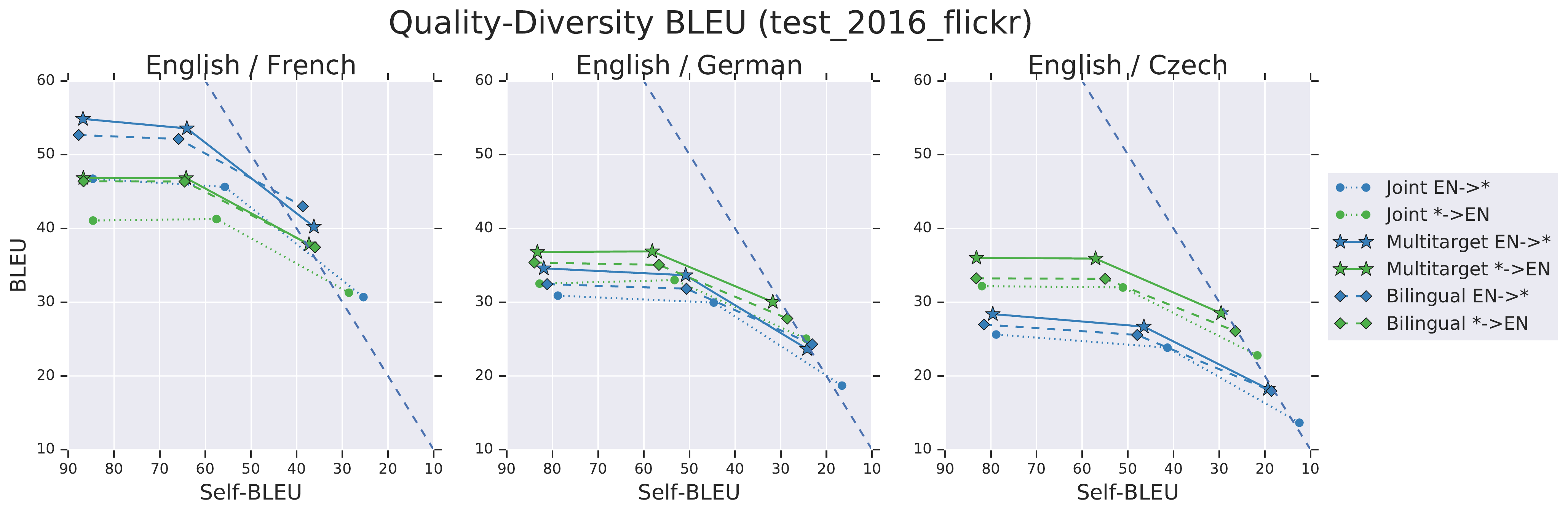}
    \caption{Quality-Diversity BLEU curve for several \modelabbv{} models (bilingual, multitarget, joint) on the Multi30k \texttt{text 2016 Flickr} test set. The dotted diagonal line signifies BLEU equals Self-BLEU. Points indicate different temperatures, from 0.1 (low diversity, left in the graph) to 1.0 (high diversity, right in the graph)}
    \label{fig:qd_cond_2016_flickr}
\end{figure*}
\begin{figure*}[t]
\small
\parbox{\textwidth}{\dotfill}
\begin{flushleft}
\vspace{-0.5em}
\textbf{English Groundtruth:}  A young boy, wearing a chef's hat and apron, is cutting sausages in a kitchen. \\
\textbf{French Groundtruth:} Un jeune garçon, portant une toque et un tablier, coupe des saucisses dans une cuisine. \\
\textbf{German Groundtruth:} Ein kleiner Junge mit Kochmütze und Schürze schneidet in einer Küche Würstchen. \\
\parbox{\textwidth}{\dotfill}
\textbf{English Seed:} \gray{A young boy,} \\
\textbf{French Seed:} \gray{portant une toque et un tablier,} \\
\textbf{German Seed:} \gray{chneidet in einer Küche Würstchen.} \\
\parbox{\textwidth}{\dotfill}
\textbf{English:} \gray{A young boy} \black{, wearing a hat} \gray{,} \black{and an apron grilling hotdogs in the kitchen.} \\
\textbf{French:} \black{Un jeune garçon} \gray{portant une toque et un tablier,} \black{faisant cuire du citron et des hotdogs dans la cuisine.} \\
\textbf{German:} \black{Ein junger Mann trägt eine Mütze und} \gray{schneidet in einer Küche Würstchen.} \\
\vspace{0.5em}
\textbf{English:} \gray{A young boy} \black{, wearing a hat and a apron, is in a kitchen} \gray{,} \black{cutting with various foods on it.} \\
\textbf{French:} \black{Un jeune garçon,} \gray{portant une toque et un tablier,} \black{est dans une cuisine en projetant des poêles de la nourriture.} \\
\textbf{German:} \black{Ein kleiner Junge mit Hut und Schürze} \gray{schneidet in einer Küche Würstchen.} \\
\vspace{0.5em}
\textbf{English:} \gray{A young boy,} \black{wearing an orange hat and apron, puts barbecue chicken in a kitchen.} \\
\textbf{French:} \black{Un jeune garçon,} \gray{portant une toque et un tablier,} \black{coupant du poulet dans une cuisine.} \\
\textbf{German:} \black{Ein kleiner Junge in einer weißen Mütze und mit Schürze} \gray{schneidet in einer Küche Würstchen} \black{glas} \gray{.} \\
\vspace{0.5em}
\textbf{English:} \gray{A young boy,} \black{wearing a blue hat and apron, is cooking meat in a kitchen.} \\
\textbf{French:} \black{Un petit garçon,} \gray{portant une toque et un tablier,} \black{fait la cuisine dans une cuisine.} \\
\textbf{German:} \black{Ein kleiner Junge mit blauer Mütze und} \gray{schneidet in einer Küche Würstchen.} \\
\end{flushleft}
\caption{Example partially conditional generation samples. The seed text is shown in gray, with several different in-filling samples from the model in black. The samples show reasonable consistency and diversity.}
\label{fig:part_cond_2016_flickr}
\end{figure*}

\subsection{Conditional Bilingual Generation: Quality-Diversity Trade-off}
We first evaluated the models on \textit{conditional} generation task by sampling bilingual translations (1 source, 1 target language) for each of the 12 language pair directions. 
We sample the token and location $(c,l) \sim p(c,l|x,\hat{y})$ from the partial canvas at each iteration, generating 100 hypothesis translations per source sentence, at softmax temperature $\tau = {0.1, 0.5, 1.0}$. At each temperature and model, we computed the \textit{quality} of the generated samples by computing the BLEU score between the reference translation and the samples, and the \textit{diversity} by computing the pairwise BLEU between the 100 samples per source, also known as Self-BLEU \cite{zhu2018texygen}. Lower Self-BLEU indicates the higher the diversity as there is less overlap between the samples. 

Figure \ref{fig:qd_cond_2016_flickr} illustrates the Quality-Diversity trade-off for the three models for different translation pairs involving English as one of the languages. The top right portion of the graph is the ideal area. We observed that the Multitarget model outperformed the Bilingual model at a lower temperature (both higher quality and diversity), and at a higher temperature, slightly above or below in quality but still higher diversity. Note that only one single Multitarget model was used for all language pair at inference time, while each bilingual model was different for each language pair curve. Therefore, a single Multitarget \modelabbv{} model could outperform specialized bilingual KERMIT models.

\subsection{Partial Conditioning Multilingual Generation}
We demonstrate our model's ability to generate infilling for partial conditioning over the multiple channels. To be explicit, we seed each channel with a few (different) words, and sample from the model. We ask the model what text completions would best fit under the model's posterior.
Figure \ref{fig:part_cond_2016_flickr} highlights several examples for (English, French, German) sentence completion. 
We took an example from the \texttt{test 2016 Flickr} test set and split it into 3 chunks--beginning in English, middle in French, and ending in German--and sample completion. 
The model can generate a set of diverse, coherent examples (Figure \ref{fig:part_cond_2016_flickr}).

\begin{figure*}[t]
    \small
    \centering
    \begin{tabular}{|c|c|p{11cm}|}
        \toprule
        \textbf{Model} & \textbf{Language} &  \textbf{Generated Sentences} \\
        \midrule
        Joint &English &  A young man in a blue jacket walking up a mountain. \\
        & French  & Un jeune homme en veste bleue descendant une paroi rocheuse en horu. \\
        & German  & Ein junger Mann in einer blauen Jacke klettert eine Felswand hoch.  \\
        & Czech   & Mladý muž v modré bundě stoupá po horách. \\
        &   & $\approx$``\textit{Young {\color{red}{men}} in blue jackets ascend and climb mountains.}" \color{green}{\checkmark}\\
        \midrule
        Biling. & English & Two small white dogs are holding the duck in a fenced yard. \\
        & French & Deux petits chiens blancs tenant un canard dans une cour clôturée. \\
        & German & Zwei kleine weiße Hunde halten eine gelbe Ente in einem eingezäunten Hof. \\
        & Czech & Dva malí chlapci drží žlutou panou venku u žlutého oploceném nádvoří. \\
        &       &$\approx$``\textit{Two little {\color{red}{boys}} holding a {\color{red}{yellow gentleman}} outside by a {\color{red}{yellow}} fenced courtyard.}" \color{red}{\ding{55}}\\
        \bottomrule
    \end{tabular}
    \caption{Example unconditional text generation samples from the Joint (top) and chain of Bilingual model (bottom). The Joint model generates one long sequence, and we split them into the four sentences in each language. In contrast, Bilingual generates a complete sentence in each language conditioned on the previous sentence above.}
    \label{fig:uncond_sampling}
\end{figure*}

\begin{figure*}[!h]
  \begin{center}
    \includegraphics[width=0.85\textwidth]{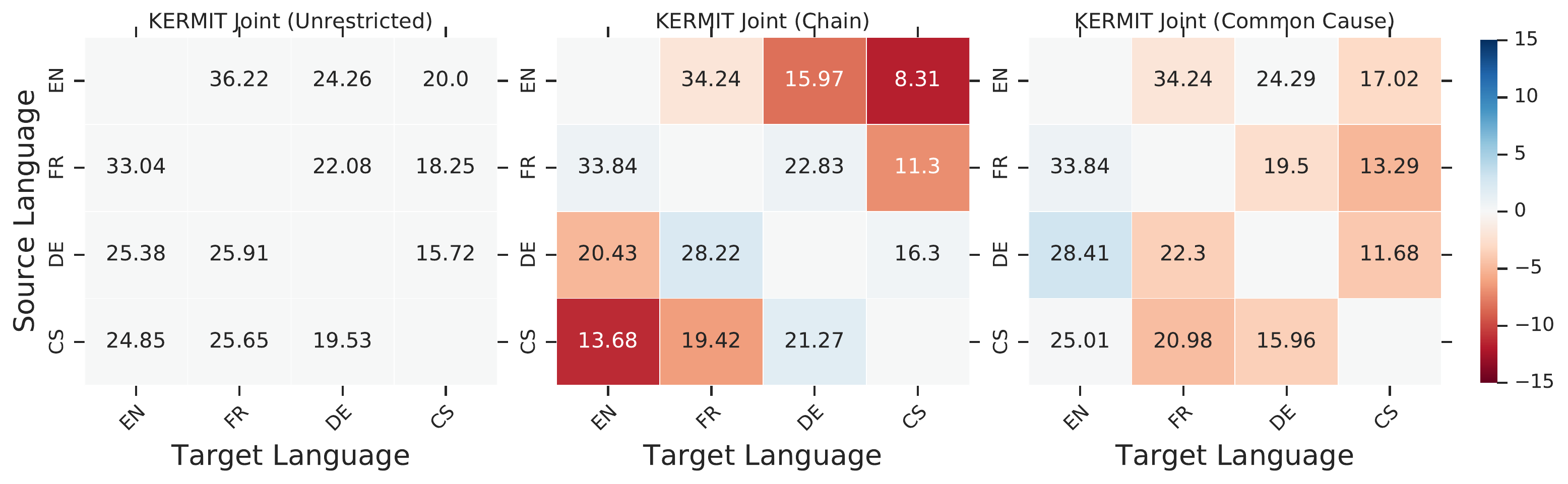}
  \end{center}
  \caption{Pseudo-Target BLEU for self-consistency for unconditional multilingual generation. Colour shading indicates the difference compared to the Joint model (unrestricted) generation.}
  \label{fig:pseudotarget_bleu}
  \vspace{-3pt}
\end{figure*}

\subsection{Unconditional Multilingual Generation}
We then evaluated the models on \textit{unconditional} multilingual generation task to generate a sentence each in all 4 languages such that they correspond to each other. For the Joint model, we perform 3 types of sampling: (1) unrestricted, (2) chain, and (3) common cause. For unrestricted, we sampled one (token, location) at each iteration starting from an empty canvas, allowing the model to insert a token in any language until all slots were marked as completed. In the chain generation, we first restrict to generating English sentence one token at a time, then sampled French, German, and Czech in order, conditioned on the last sentence in the previous language. For common cause, we reuse the same English and French sampled sentences and generate the German and Czech conditioned on the English sentence (i.e., 3 languages are all conditioned on English). 

Given these sets of sentences in 4 languages, for each pair of language direction, we computed a \textit{pseudo} target by using a separately trained (on Multi30k) vanilla Transformer \citep{vaswani-nips-2017} and performed beam search (size 5) to translate the chosen source language sample. Figure \ref{fig:pseudotarget_bleu} visualizes the pseudo target BLEU score for different source-target language pairs when comparing the Joint model under different types of sampling. The shaded colour represents the difference between the current sampling scheme versus the unrestricted reference. We observe that letting the model sample in unrestricted order was better than either the chain or the common cause sampling. 
\section{Conclusion and Future Work}
In this paper, we presented the Multichannel Generative Language Model (MGLM). 
MGLM is a generative joint distribution model that marginalizes over all possible factorizations within and across channels.
MGLM endows flexible inference, including unconditional, conditional, and partially observed generation. 
We experimented with those inference modes using the Multi30K dataset containing English, French, Czech, and German.
We provide qualitative samples sampled unconditionally from the generative joint distribution.
We also quantitatively analyze the quality-diversity trade-offs and find MGLM outperform traditional bilingual discriminative models.

Our work focused on a specific instantiation of channels as languages. 
However, MGLM is not limited to only languages and can generalize to other notions of channels. 
In future work, we will consider other textual channels, such as paraphrases, premises and hypotheses, questions and answers, and multimodal channels, such as images.
Another direction can investigate scaling MGLM to dozens/hundreds of channels.
Fully generative models still often lag behind purely discriminative counterparts in performance, but we hope our work motivates future research on building generative joint distribution models of the world.

\section*{Acknowledgement}
We give thanks to Mohammad Norouzi, Lala Li, Sara Sabour, Geoffrey Hinton, Silviu Pitis, and the Google Brain team for useful discussions and feedbacks. 

\bibliographystyle{acl_natbib}
\bibliography{mglm}

\appendix
\section{Appendices}\label{sec:appendix}
\subsection{Multi30k Dataset Description} \label{app:dataset}
The statistics of the Multi30K dataset (Task 1) are summarized in Table \ref{tab:multi30k_stats}. The average number of words across training, validation, and 2016 test for English is 11.9, and for German is 11.1 \citep{elliott-acl-2016}. Since we use SentencePiece \citep{kudo2018sentencepiece}, \modelabbv{} sees more number of tokens per sentence on average. 

\begin{table}[!h]
\centering
\small
\begin{tabular}{lc}
\toprule
\bfseries Subset  & Number of Sentences \\
\midrule
Training & 29,000 \\
Validation & 1,014 \\
\texttt{Test 2016 Flickr} & 1,000 \\
\texttt{Test 2017 Flickr} & 1,000 \\
\texttt{test 2017 MSCOCO} & 461 \\
\bottomrule
\end{tabular}%
\caption{Multi30k English $\rightarrow$ Czech test BLEU.}
\label{tab:multi30k_stats}
\end{table}

\subsection{Additional Quality-Diversity Curves For Conditional Generation} \label{app:qd-curve}
\begin{figure*}[t]
    \centering
    \includegraphics[width=\textwidth]{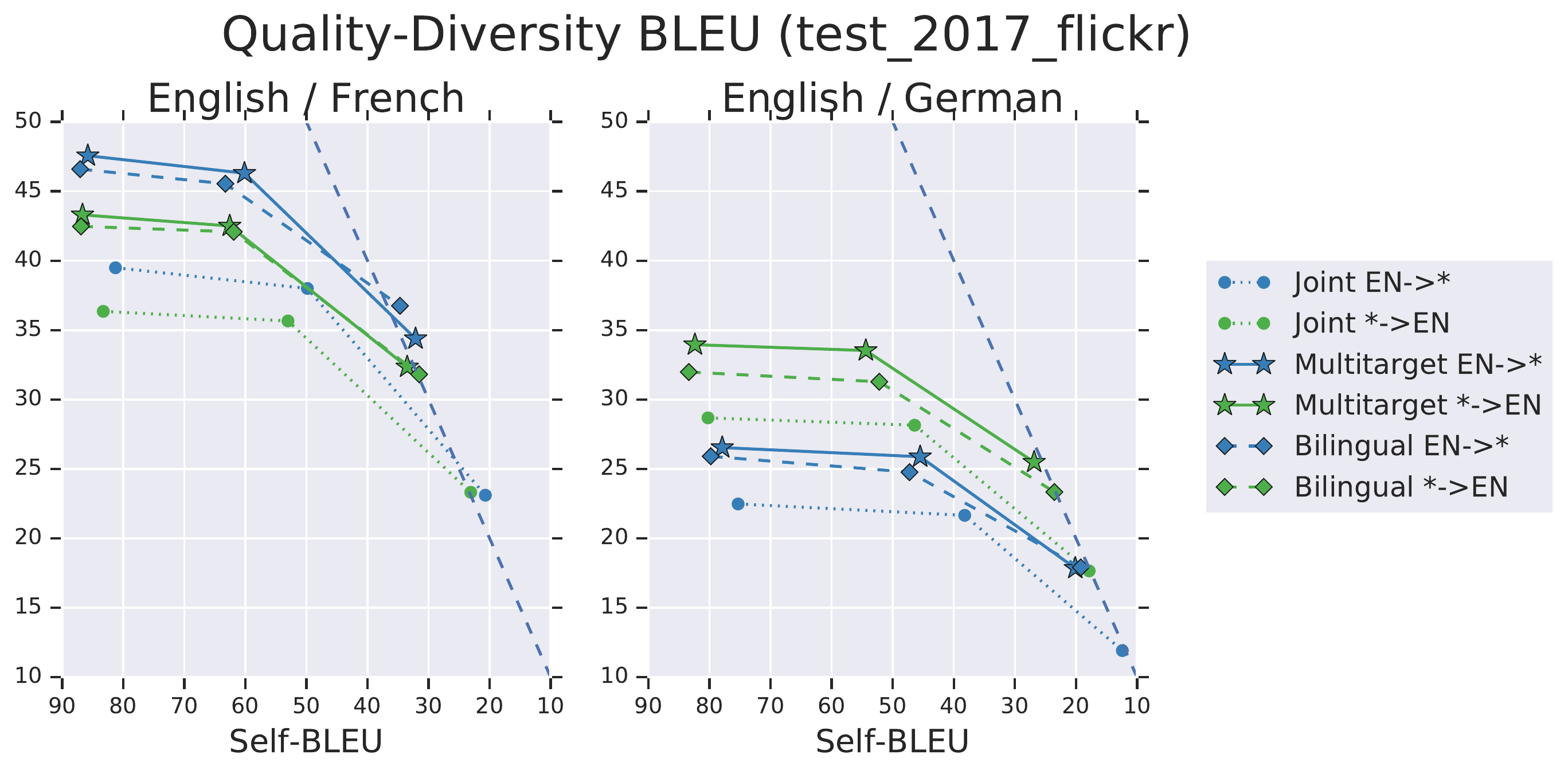}
    \caption{Quality-Diversity BLEU curve for several \modelabbv{} models (bilingual, multitarget, joint) on the Multi30k \texttt{text 2017 Flickr} test set. The dotted diagonal line signifies BLEU equals Self-BLEU. Points indicate different temperatures, from 0.1 (low diversity, left in the graph) to 1.0 (high diversity, right in the graph)}
    \label{fig:qd_cond_2017_flickr}
\end{figure*}

\begin{figure*}[t]
    \centering
    \includegraphics[width=\textwidth]{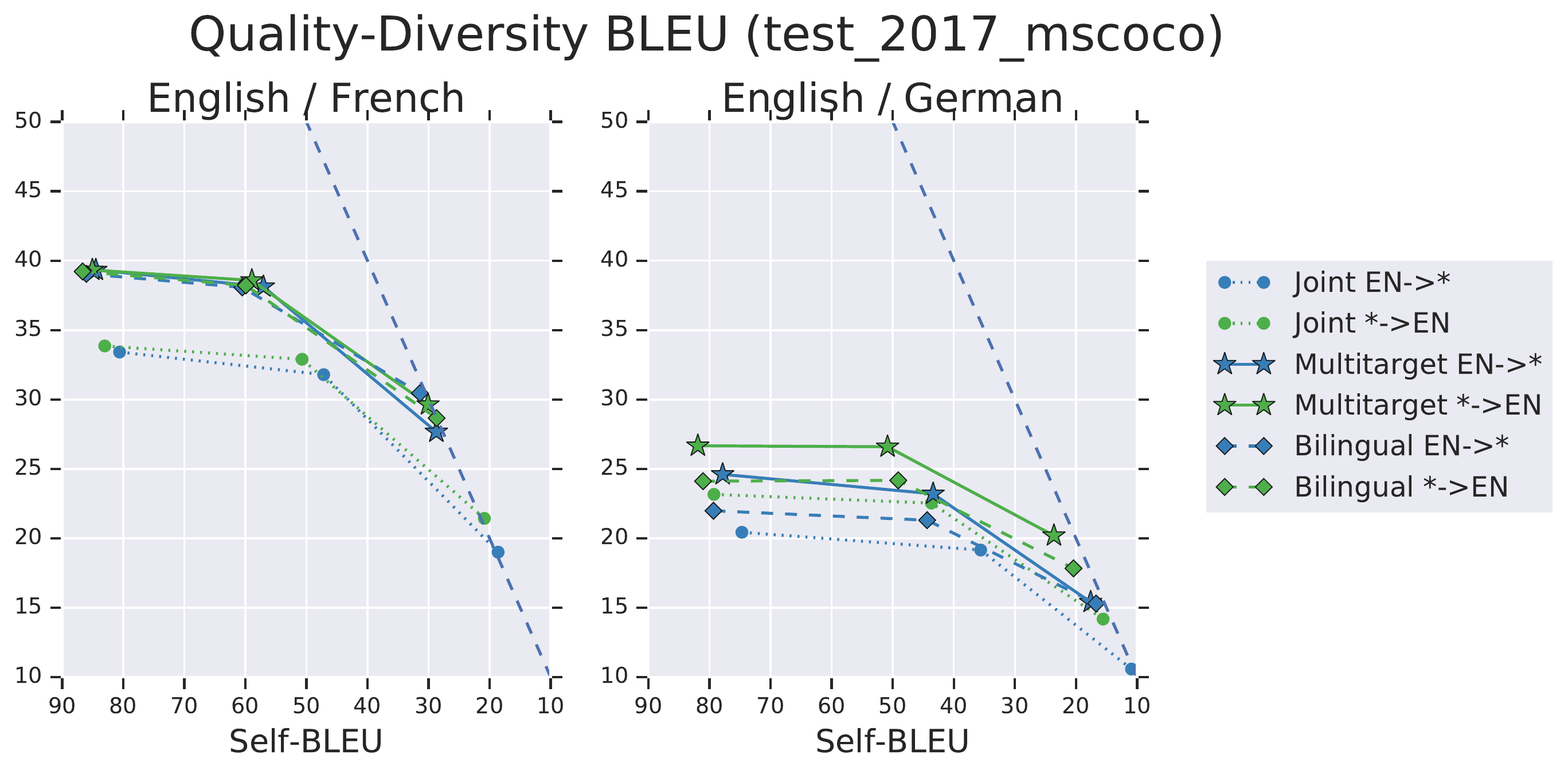}
    \caption{Quality-Diversity BLEU curve for several \modelabbv{} models (bilingual, multitarget, joint) on the Multi30k \texttt{text 2017 MSCOCO} test set. The dotted diagonal line signifies BLEU equals Self-BLEU. Points indicate different temperatures, from 0.1 (low diversity, left in the graph) to 1.0 (high diversity, right in the graph)}
    \label{fig:qd_cond_2017_mscoco}
\end{figure*}

We include additional Quality-Diversity Curves For Conditional Generation: Figure \ref{fig:qd_cond_2017_flickr} for the \texttt{test 2017 Flickr}, and Figure \ref{fig:qd_cond_2017_mscoco} for the \texttt{test 2017 MSCOCO}. 

\subsection{Additional Multi30K Translation Results} \label{app:more_multi30k_bleu}
We include additional Multi30K Translation Results: Table \ref{tab:multi30k_en2fr} for English to French, Table \ref{tab:multi30k_en2cs} for English to Czech, and Table \ref{tab:multi30k_de2en} for German to English. 

\subsection{Unconditional Sampling Generation} \label{app:uncon_samp_gen}
Figure \ref{fig:uncond_sampling_process} illustrates the serial sampling (one token at a time) from the joint model, every 20 timesteps.

\FloatBarrier
\begin{table*}[!h]
\centering
\small
\begin{tabular}{lcccc}
\toprule
\bfseries Model & Inference & Test2016  & Test2017 & MSCOCO  \\
\midrule
Bilingual (EN $\rightarrow$ FR) & EN $\rightarrow$ FR & 58.80 & 50.35  & \textbf{42.82}  \\
Bilingual (EN $\leftrightarrow$ FR) & EN $\rightarrow$ FR & \textbf{59.29} & \textbf{52.13} & 42.17  \\
\midrule
Multi-target (EN $\rightarrow$ Rest) & EN $\rightarrow$ FR & 58.08 & 50.39 & 42.19 \\
                                     & EN $\rightarrow$ \textbf{FR},CS,DE & 58.52 & 50.49 & 41.53 \\
\midrule
Multi-target (Any $\rightarrow$ Rest) & EN $\rightarrow$ FR & 57.64 & 50.01 & 40.18 \\
                                     & EN $\rightarrow$ \textbf{FR},CS,DE & 57.35 & 48.13 & 39.98  \\
\midrule
Joint ($p(EN,FR,CS,DE)$) & EN $\rightarrow$ FR & 50.87 & 40.69 & 33.93 \\
                        & EN $\rightarrow$ \textbf{FR},CS,DE & 48.85 & 39.92 & 33.45  \\
\bottomrule
\end{tabular}%
\caption{Multi30k English $\rightarrow$ French test BLEU.}
\label{tab:multi30k_en2fr}
\end{table*}

\begin{table*}[h]
\centering
\small
\begin{tabular}{lcc}
\toprule
\bfseries Model & Inference & Test2016   \\
\midrule
Bilingual (EN $\rightarrow$ CS) & EN $\rightarrow$ CS & 28.58 \\
Bilingual (EN $\leftrightarrow$ CS) & EN $\rightarrow$ CS & 29.03  \\
\midrule
Multi-target (EN $\rightarrow$ Rest) & EN $\rightarrow$ CS  & \textbf{30.48}  \\
                                     & EN $\rightarrow$ FR,\textbf{CS},DE  & 30.15 \\
\midrule
Multi-target (Any $\rightarrow$ Rest) & EN $\rightarrow$ CS &  30.11 \\
                                     & EN $\rightarrow$ FR,\textbf{CS},DE  & 30.11\\
\midrule
Joint ($p(EN,FR,CS,DE)$) & EN $\rightarrow$ CS  & 26.45  \\
                        & EN $\rightarrow$ FR,\textbf{CS},DE  & 26.35   \\
\bottomrule
\end{tabular}%
\caption{Multi30k English $\rightarrow$ Czech test BLEU.}
\label{tab:multi30k_en2cs}
\end{table*}

\begin{table*}[h!]
\centering
\small
\begin{tabular}{lcccc}
\toprule
\bfseries Model & Inference & Test2016  & Test2017 & MSCOCO  \\
\midrule
Bilingual (DE $\rightarrow$ EN) & DE $\rightarrow$ EN & 39.40 & 34.90 & 27.75 \\
Bilingual (EN $\leftrightarrow$ DE) & DE $\rightarrow$ EN & 40.52 & 35.66 & 28.61  \\
\midrule
Multi-target (DE $\rightarrow$ Rest) & DE $\rightarrow$ EN & \textbf{40.75} & 36.38 & \textbf{28.91} \\
                                     & DE $\rightarrow$ \textbf{EN}, FR,CS & 39.72 & 35.95 & 28.20 \\
\midrule
Multi-target (Any $\rightarrow$ Rest) & DE $\rightarrow$ EN & 40.69 & 36.02 & 28.89\\
                                     & DE $\rightarrow$ \textbf{EN}, FR,CS & 39.97 & \textbf{37.07} & 28.62  \\
\midrule
Joint ($p(EN,FR,CS,DE)$) & DE $\rightarrow$ EN & 38.44 & 30.82 & 25.46 \\
                        & DE $\rightarrow$ \textbf{EN}, FR,CS & 36.30 & 29.68 & 24.87 \\
\bottomrule
\end{tabular}%
\caption{Multi30k German $\rightarrow$ English test BLEU.}
\label{tab:multi30k_de2en}
\end{table*}

\begin{figure*}[h]
    \small
    \centering
    \begin{tabular}{|c|c|p{10cm}|}
        \toprule
        \textbf{Iterations} & \textbf{Language} &  \textbf{Generated Sentence from Joint Model} \\
        \midrule
        1 & English &  \\
         & French  &  \\
         & Czech   & Mladý \\
         & German  &  \\
        \midrule
        20 & English & \\
        & French  & descendant\\ 
        & Czech   &  Mladý muž v modré bundě stoupá po\\
        & German  &  Mann klettert. \\
        \midrule
        40 & English & blue jacket walking up a mountain.\\
        & French  & veste descendant paroi rocheuse en \\
        & Czech   &  Mladý muž v modré bundě stoupá po horách. \\
        & German  &  Mann klettert. \\
        \midrule
        60 & English & A man blue jacket walking up a mountain. \\
        & French  &  veste bleue descendant une paroi rocheuse en horu.\\
        & Czech   &  Mladý muž v modré bundě stoupá po horách.\\
        & German  &  Mann einer blauen klettert eine hoch.\\
        \midrule
        80 & English & A young man in blue jacket walking up a mountain.\\
        & French  & veste bleue descendant une paroi rocheuse en horu.\\
        & Czech   &  Mladý muž v modré bundě stoupá po horách.\\
        & German  &  Ein junger Mann in einer blauen Jacke klettert eine Felswand hoch.\\
        \midrule
        96 & English &  A young man in a blue jacket walking up a mountain. \\
        & French  & Un jeune homme en veste bleue descendant une paroi rocheuse en horu. \\
        & Czech   & Mladý muž v modré bundě stoupá po horách. \\
        & German  & Ein junger Mann in einer blauen Jacke klettert eine Felswand hoch.  \\
        \bottomrule
    \end{tabular}
    \caption{Example of serial sampling unconditional text generation from the joint $p(EN, FR, CS, DE)$ model, over 96 insertion time steps. Note that the model generates one long sequence, and we split them into the resulting four sentences in each language here.}
    \label{fig:uncond_sampling_process}
\end{figure*}

\end{document}